\documentclass[runningheads]{llncs}
\usepackage[T1]{fontenc}
\usepackage{graphicx,verbatim}
\usepackage{booktabs}
\usepackage{multirow}
\usepackage{amsmath}
\usepackage{amssymb}
\usepackage[table,xcdraw]{xcolor}
\usepackage{xcolor}

\newcommand{\cmark}{\ding{51}}  
\usepackage{pifont}

\begin{document}
\title{Segmentation-before-Staining Improves Structural Fidelity in Virtual IHC-to-Multiplex IF Translation}
\titlerunning{Segmentation-before-Staining for Virtual IHC-to-mIF Translation}

\author{Junhyeok Lee$^*$\inst{1} \and
Han Jang$^*$\inst{2} \and
Heeseong Eom\inst{1} \and
Joon Jang\inst{3} \and
Kyu Sung Choi\inst{4,5,6}}
\authorrunning{J. Lee et al.}
\institute{Interdisciplinary Program in Cancer Biology, Seoul National University College of Medicine, Seoul, Republic of Korea \and
Interdisciplinary Program in Bioengineering, Seoul National University, Seoul, Republic of Korea \and
Department of Biomedical Sciences, Seoul National University, Seoul, Republic of Korea \and
Department of Radiology, Seoul National University Hospital, Seoul, Republic of Korea \and
Department of Radiology, Seoul National University College of Medicine, Seoul, Republic of Korea \and
Healthcare AI Research Institute, Seoul National University Hospital, Seoul, Republic of Korea\\
\email{\{jhlee0619, hanjang, seong6466, jangjoon7\}@snu.ac.kr, ent1127@snu.ac.kr}}


\maketitle
\renewcommand{\thefootnote}{}
\footnotetext{$^*$ Equal contribution.}
\renewcommand{\thefootnote}{\arabic{footnote}}

\begin{abstract}

Multiplex immunofluorescence~(mIF) enables simultaneous single-cell quantification of multiple biomarkers within intact tissue architecture, yet its high reagent cost, multi-round staining protocols, and need for specialized imaging platforms limit routine clinical adoption.
Virtual staining can synthesize mIF channels from widely available brightfield immunohistochemistry~(IHC), but current translators optimize pixel-level fidelity without explicitly constraining nuclear morphology.
In pathology, this gap is clinically consequential: subtle distortions in nuclei count, shape, or spatial arrangement propagate directly to quantification endpoints such as the Ki67 proliferation index, where errors of a few percent can shift treatment-relevant risk categories.
This work introduces a supervision-free, architecture-agnostic conditioning strategy that injects a continuous cell probability map from a pretrained nuclei segmentation foundation model as an explicit input prior, together with a variance-preserving regularization term that matches local intensity statistics to maintain cell-level heterogeneity in synthesized fluorescence channels.
The soft prior retains gradient-level boundary information lost by binary thresholding, providing a richer conditioning signal without task-specific tuning.
Controlled experiments across Pix2Pix with U-Net and ResNet generators, deterministic regression U-Net, and conditional diffusion on two independent datasets demonstrate consistent improvements in nuclei count fidelity and perceptual quality, as the sole modifications.
Code will be made publicly available upon acceptance.


\end{abstract}

\section{Introduction}

Multiplex immunofluorescence~(mIF) enables simultaneous detection of multiple protein markers within a single tissue section, supporting quantitative phenotyping of immune infiltrates and spatial biomarker analysis that informs prognosis and immunotherapy response~\cite{bollhagen2024highly,tan2020overview,hoyt2021multiplex}.
However, mIF requires multi-round tyramide signal amplification, spectral unmixing, and dedicated imaging platforms, with per-slide costs and turnaround times far exceeding those of conventional immunohistochemistry~(IHC)~\cite{harms2023multiplex,sun2025exploring}.
Brightfield IHC is inexpensive and universally available, yet limited to one or two markers per section and reliant on manual scoring by pathologists, a labor-intensive process with substantial inter-observer variability~\cite{rimm2023pathologists}.
This is especially problematic for proliferation markers such as Ki67, whose positive fraction serves as a key prognostic indicator across multiple cancer types and directly influences grading, molecular subtyping, and adjuvant treatment decisions~\cite{nielsen2021assessment,yerushalmi2010ki67}.
Virtual staining bridges this gap by translating IHC into multi-channel mIF representations~\cite{ghahremani2022deep,ghahremani2022deepliifui,bai2023deep,rivenson2019virtual}.

Several architectures have been proposed for virtual staining, including multitask GANs~\cite{ghahremani2022deep}, multi-scale supervision~\cite{liu2022bci}, task-specific scoring losses~\cite{peng2024advancing}, weakly-supervised feature extraction~\cite{li2024virtual}, and generative multiplexing from H\&E~\cite{pati2024accelerating}.
While these methods achieve high perceptual fidelity, standard metrics can mask clinically relevant structural errors~\cite{klockner2025h}.
Adversarial generators may alter nuclei boundaries or merge adjacent cells~\cite{isola2017pix2pix}, and stochastic samplers can introduce morphological drift~\cite{ho2020ddpm}.
Such distortions can bias the Ki67 positive fraction sufficiently to cross clinical decision thresholds~\cite{yerushalmi2010ki67,boyaci2025global}.

This observation motivates a simple hypothesis that providing nuclei-derived structure as an explicit input prior before translation improves structural preservation and downstream quantification, even under paired pixel-level supervision.
The proposed approach leverages Cellpose~\cite{stringer2021cellpose}, a generalist cell segmentation foundation model pretrained on diverse microscopy modalities including fluorescence, brightfield, and histopathology images, to extract a continuous per-pixel cell probability map directly from the IHC input without any domain-specific fine-tuning.
This soft representation is concatenated to the IHC input as a single additional channel, providing the translator with gradient-level boundary information that binary masks discard through hard thresholding.
The training loss is further augmented with a variance-preserving regularization term that matches local second-order statistics between synthesized and ground-truth channels, preserving the cell-level intensity heterogeneity that regression objectives tend to smooth.

The contributions are as follows:
(i)~a supervision-free, architecture-agnostic soft prior conditioning requiring no task-specific training or threshold selection;
(ii)~a variance-preserving loss that maintains local intensity heterogeneity in synthesized fluorescence channels;
(iii)~controlled benchmarking across Pix2Pix with U-Net and ResNet generators, regression U-Net, and conditional diffusion on two independent datasets;
and (iv)~evaluation via nuclei count fidelity and Ki67 quantification error linking synthesis quality to diagnostic utility.

\begin{figure}[t]
\centering
\includegraphics[width=\linewidth]{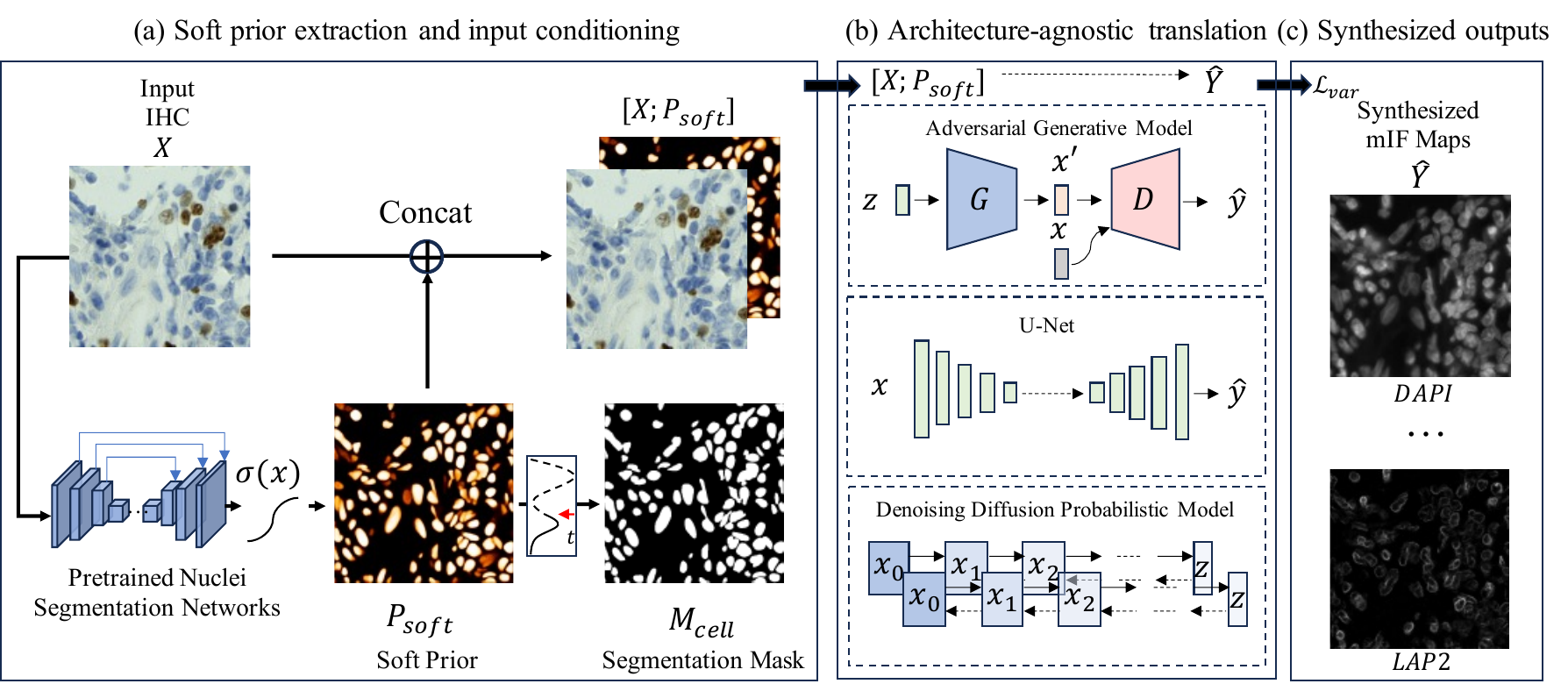}
\caption{
\textbf{Overview of the proposed soft prior conditioning framework.}
(a)~Cellpose~\cite{stringer2021cellpose} extracts a continuous probability map $P_\text{soft}$, concatenated as a fourth channel.
(b)~The conditioned input is passed to any translation architecture.
(c)~Multi-channel outputs; $\mathcal{L}_\text{var}$ is added to the base objective to constrain local intensity statistics.
}
\label{fig:overview}
\end{figure}

\section{Method}

The proposed approach extends standard paired IHC-to-mIF translation with two components, both architecture-agnostic and requiring no additional training or paired annotations, as illustrated in Fig.~\ref{fig:overview}.

\subsection{Soft Structural Conditioning for Paired IHC-to-mIF Translation}
The task of improving structural reliability in virtual staining is formulated as an input-level conditioning problem rather than an architectural redesign.
Given paired, co-registered IHC RGB patches ($X \in \mathbb{R}^{H \times W \times 3}$) and their corresponding mIF stacks ($Y \in \mathbb{R}^{H \times W \times K}$) containing a nuclear reference channel, the goal is to synthesize multi-channel mIF targets while preserving fine-grained nuclear morphology and spatial arrangement.

The key idea is to augment the IHC input with a continuous cell probability map extracted from a pretrained nuclei segmentation model, providing the translator with a soft structural prior that encodes per-pixel nuclei likelihood without imposing hard binarization decisions.
This formulation leaves the translation architecture and base training objective untouched, with only the input tensor modified, as illustrated in Fig.~\ref{fig:overview}.
The translation model $G$ generates the predicted mIF stack $\hat{Y}$ as:
\begin{equation}
\hat{Y} = G([X;\, P_{\text{soft}}]),
\end{equation}
where $[\cdot\,;\,\cdot]$ denotes channel-wise concatenation of the 3-channel RGB input and the single-channel soft probability map (4 channels total).

\subsection{Soft Prior Generation}
The soft prior is extracted from a single forward pass of Cellpose~\cite{stringer2021cellpose}, a pretrained generalist cell segmentation foundation model, applied directly to the brightfield IHC input as shown in Fig.~\ref{fig:overview}a.
No task-specific training or access to paired mIF data is required, ensuring full decoupling from the translation pipeline.

Specifically, Cellpose produces a per-pixel cell probability output $P_{\text{soft}} \in [0,1]^{H \times W}$ via a sigmoid activation prior to the binarization step used for instance mask generation.
This continuous representation retains gradient-level information about cell boundaries, as pixels near nuclei peripheries receive intermediate probabilities rather than being forced to a hard $\{0,1\}$ decision. Sub-threshold structures such as partial nuclei at patch borders, mitotic figures with diffuse chromatin, or overlapping cells are likewise preserved rather than discarded.

Compared with a binarized mask, the soft map is an information-theoretic superset because the binary mask is recoverable via thresholding but not vice versa. It also eliminates the threshold hyperparameter and propagates smoother gradients to early encoder layers.
The binarized counterpart $M_\text{cell} = 1[P_\text{soft} > t]$ is retained only for ablation comparison.

\subsection{Variance-Preserving Regularization}
\label{sec:var_loss}
Standard pixel-level objectives ($\ell_1$/$\ell_2$) penalize mean intensity error but do not explicitly constrain local texture statistics.
Regression-based models tend to smooth high-frequency intensity patterns in fluorescence channels, reducing local contrast and undermining downstream cell detection~\cite{johnson2016perceptual}.

To preserve cell-level intensity heterogeneity, the base translation loss is augmented with a local variance matching term.
The local variance map $V_k(\cdot)$ is defined as:

\begin{equation}
V_k(I) = \mu_k(I^{2}) - \bigl[\mu_k(I)\bigr]^{2},
\end{equation}

where $\mu_k(\cdot)$ denotes average pooling with kernel size $k$ and reflect padding, and ${2}$ is the element-wise square.
The variance-preserving loss then penalizes discrepancies in local second-order statistics:
\begin{equation}
\mathcal{L}_{\text{var}} = \bigl\| V_k(\hat{Y}) - V_k(Y) \bigr\|_2^2.
\end{equation}

The total training objective, illustrated in Fig.~\ref{fig:overview}c, becomes:
\begin{equation}
\mathcal{L}_{\text{total}} = \mathcal{L}_{\text{base}} + \lambda_{\text{var}}\, \mathcal{L}_{\text{var}},
\end{equation}
where $\mathcal{L}_{\text{base}}$ denotes the architecture-specific loss.
For adversarial models, $\mathcal{L}_{\text{base}} = \mathcal{L}_{\text{GAN}} + \lambda_{\ell_1}\mathcal{L}_{\ell_1}$, and for deterministic regression, $\mathcal{L}_{\text{base}} = \mathcal{L}_{\ell_1}$.

\subsection{Clinical Fidelity Metrics}
\label{sec:metrics}
To assess whether synthesized mIF channels support reliable downstream quantification beyond standard image-quality measures, two domain-specific metrics are introduced.
All thresholds are determined on the validation split and fixed at test time.

\paragraph{Ki67 positivity error.}
Given nuclei instances $\{c_i\}_{i=1}^{N}$ detected from the ground-truth nuclear reference channel, each nucleus is classified as Ki67-positive when its mean synthesized marker intensity exceeds a validation-selected threshold $\tau$.
The Ki67 positive fraction is defined as:
\begin{equation}
f^{+} = \frac{1}{N}\sum_{i=1}^{N} \mathbf{1}[\bar{I}(c_i) > \tau],
\end{equation}
and the metric is the absolute error $|f^{+}_{\text{pred}} - f^{+}_{\text{gt}}|$, averaged across test images.

\paragraph{Per-pixel MAE.}
For HNSCC, where only the DAPI nuclear channel is translated, the per-pixel mean absolute error $\frac{1}{HW}\|\hat{Y} - Y\|_1$ is reported as a direct measure of reconstruction accuracy.

\section{Experiments and Results}
\label{sec:experiments}

\subsection{Experimental Setup}

\paragraph{Datasets.}
Two publicly available paired datasets are used.
DeepLIIF~\cite{ghahremani2022deepliifui} provides co-registered IHC and 3-channel mIF stacks (DAPI, Lap2, Ki67) from multiple organ sites stained with DAB chromogen ($n_\text{train}{=}2{,}396$, $n_\text{test}{=}598$).
HNSCC~\cite{ghahremani2023ai} provides co-registered mIF and mIHC stacks with nuclear, immune, and tumor markers from head-and-neck squamous cell carcinoma ($n_\text{train}{=}197$, $n_\text{test}{=}71$, AEC chromogen).
The DAPI channel is selected to maintain a shared target with DeepLIIF.
Patches are resized to $256{\times}256$ and normalized to $[-1,1]$.
All splits are enforced at the case level to prevent leakage across correlated patches.

\begin{table}[t]
\caption{%
Translation quality with off-the-shelf nuclei prior.
+Prior adds the soft probability map and variance-preserving loss.
DeepLIIF~\cite{ghahremani2022deepliifui}  translates IHC to 3-channel mIF.
HNSCC~\cite{ghahremani2023ai} translates mIHC to DAPI.
$^\dagger$Ki67 positivity error.
$^\ddagger$Per-pixel MAE on DAPI channel.
Best per architecture in \textbf{bold}.%
}
\label{tab:main}
\centering
\small
\setlength{\tabcolsep}{3pt}
\begin{tabular}{@{}ll ccc c ccc@{}}
\toprule
& & \multicolumn{3}{c}{DeepLIIF~\cite{ghahremani2022deepliifui} (3-ch mIF)}
& & \multicolumn{3}{c}{HNSCC~\cite{ghahremani2023ai} (DAPI 1-ch)} \\
\cmidrule(lr){3-5} \cmidrule(l){7-9}
Model & Config
  & SSIM\,$\uparrow$ & LPIPS\,$\downarrow$ & Ki67$^\dagger$\,$\downarrow$
  &
  & SSIM\,$\uparrow$ & LPIPS\,$\downarrow$ & pMAE$^\ddagger$\,$\downarrow$ \\
\midrule
Pix2Pix~\cite{isola2017pix2pix}  & Baseline
  & .719 & .318 & .018
 && .834 & .134 & .019 \\
(UNet) & +Prior
  & \textbf{.732} & \textbf{.296} & .018
 && \textbf{.855} & \textbf{.113} & \textbf{.017} \\
\addlinespace
Pix2Pix~\cite{isola2017pix2pix}  & Baseline
  & .759 & .293 & \textbf{.019}
 && \textbf{.880} & \textbf{.121} & .016 \\
(ResNet) & +Prior
  & \textbf{.763} & \textbf{.291} & .021
 && \textbf{.880} & .131 & \textbf{.015} \\
\addlinespace
Regression~\cite{ronneberger2015unet} & Baseline
  & .724 & .315 & .018
 && .812 & .197 & .020 \\
(UNet) & +Prior
  & \textbf{.731} & \textbf{.306} & .018
 && \textbf{.856} & \textbf{.113} & \textbf{.018} \\
\addlinespace
Diffusion~\cite{ho2020ddpm} & Baseline
  & .202 & .607 & \textbf{.052}
 && .697 & .171 & .037 \\
(DDPM) & +Prior
  & \textbf{.552} & \textbf{.394} & .066
 && \textbf{.819} & \textbf{.135} & \textbf{.021} \\
\bottomrule
\end{tabular}
\end{table}

\paragraph{Translation architectures.}
To demonstrate architecture-agnostic applicability, a controlled comparison is conducted across three representative paradigms.
The only difference between baseline and proposed variants is the concatenation of the soft prior channel and, where applicable, the variance-preserving term.
Pix2Pix~\cite{isola2017pix2pix} with a PatchGAN discriminator and $\ell_1$ reconstruction is trained using both U-Net~\cite{ronneberger2015unet} and ResNet~\cite{he2016deep} generators, with the input channel count adjusted from three to four.
Regression U-Net~\cite{ronneberger2015unet} trained with $\ell_1$ loss serves as a non-adversarial baseline, with the encoder-decoder structure otherwise identical.
Conditional denoising diffusion probabilistic model (DDPM)~\cite{ho2020ddpm} conditions the denoiser on the optionally augmented input at each step, with the noise schedule and training objective identical across configurations.

\paragraph{Training details.}
Pix2Pix models are trained with Adam ($\beta_1{=}0.5$, $\beta_2{=}0.999$) at learning rate $2{\times}10^{-4}$ for 1,000 epochs with $\lambda_{\ell_1}{=}100$.
Regression U-Net uses Adam at learning rate $10^{-4}$ for 150 epochs.
Diffusion uses a linear noise schedule with $T{=}1000$ steps, trained with Adam at learning rate $2{\times}10^{-4}$ for 1,000 epochs.
For Pix2Pix and Regression U-Net, the variance-preserving loss (Sec.~\ref{sec:var_loss}) is added with $\lambda_\text{var}{=}50$ and kernel size $k{=}15$.
All models are trained with batch size 16 and early stopping on validation loss, with 20\% of the training set held out for validation.
Experiments are conducted on two NVIDIA RTX 3090 GPUs using distributed data parallel training.

\paragraph{Evaluation metrics.}
Image-level fidelity is measured by structural similarity index (SSIM)~\cite{wang2004ssim} and learned perceptual image patch similarity (LPIPS)~\cite{zhang2018lpips} across all synthesized channels.
Two additional domain-specific metrics, Ki67 positivity error and per-pixel MAE, are defined in Sec.~\ref{sec:metrics}.

\begin{table}[t]
\caption{%
Ablation on prior representation and loss (HNSCC~\cite{ghahremani2023ai}, DAPI, $n_\text{test}{=}71$).
$M_\text{cell}$: binary mask (1\,ch).
$P_\text{soft}$: soft probability map (1\,ch).
$\mathcal{L}_\text{var}$: variance-preserving loss ($\lambda{=}50$, $k{=}15$).
Best per architecture in \textbf{bold}.
}
\label{tab:ablation}
\centering
\small
\setlength{\tabcolsep}{3pt}
\begin{tabular}{@{}ll ccc ccc@{}}
\toprule
Model & Prior
  & $M_\text{cell}$ & $P_\text{soft}$ & $\mathcal{L}_\text{var}$
  & SSIM\,$\uparrow$ & LPIPS\,$\downarrow$ & pMAE\,$\downarrow$ \\
\midrule
Pix2Pix~\cite{isola2017pix2pix}   & None       &        &        &        & .834 & .134 & .019 \\
(UNet)    & Binary     & \cmark &        &        & .846 & .175 & .017 \\
          & Soft       &        & \cmark &        & .855 & .113 & \textbf{.017} \\
          & Soft+Var   &        & \cmark & \cmark & \textbf{.855} & \textbf{.113} & .017 \\
\addlinespace
Pix2Pix~\cite{isola2017pix2pix}   & None       &        &        &        & .880 & .121 & .016 \\
(ResNet)  & Binary     & \cmark &        &        & .879 & .125 & .015 \\
          & Soft       &        & \cmark &        & \textbf{.881} & \textbf{.121} & .015 \\
          & Soft+Var   &        & \cmark & \cmark & .880 & .131 & \textbf{.015} \\
\addlinespace
Regression~\cite{ronneberger2015unet} & None      &        &        &        & .812 & .197 & .020 \\
(UNet)     & Binary    & \cmark &        &        & .840 & .149 & .018 \\
           & Soft      &        & \cmark &        & \textbf{.858} & .115 & \textbf{.017} \\
           & Soft+Var  &        & \cmark & \cmark & .856 & \textbf{.113} & .018 \\
\addlinespace
Diffusion~\cite{ho2020ddpm}  & None      &        &        &        & .697 & .171 & .037 \\
(DDPM)     & Binary    & \cmark &        &        & .755 & .165 & .028 \\
           & Soft      &        & \cmark &        & .790 & \textbf{.128} & .028 \\
           & Soft+Var  &        & \cmark & \cmark & \textbf{.819} & .135 & \textbf{.021} \\
\bottomrule
\end{tabular}
\end{table}

\begin{figure*}[t]
\centering
\includegraphics[width=0.9\textwidth]{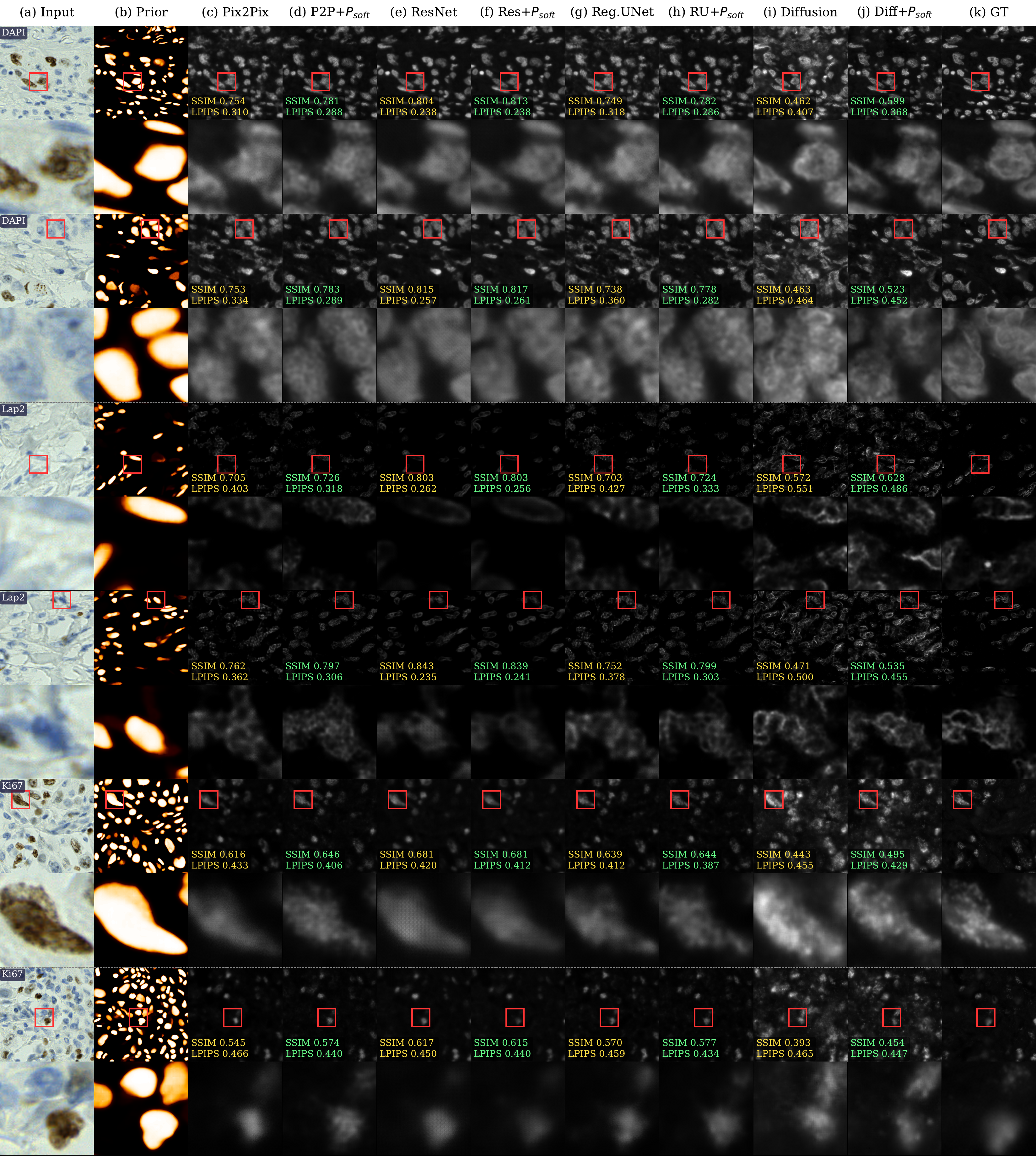}
\caption{
Qualitative comparison on DeepLIIF test cases across three target channels (DAPI, Lap2, Ki67) from bladder and lung tissue.
}
\label{fig:qual}
\end{figure*}

\subsection{Main Results}

Table~\ref{tab:main} summarizes quantitative comparisons across both datasets.
The proposed method consistently improves SSIM and LPIPS across most architectures.
On DeepLIIF, Pix2Pix (UNet) improves from .719 to .732 in SSIM and .318 to .296 in LPIPS, and Diffusion recovers from .202 to .552 in SSIM, demonstrating consistent gains across both chromogen types.
On HNSCC, Regression U-Net achieves the largest gain (SSIM .812 to .856, LPIPS .197 to .113), while Diffusion shows the most dramatic recovery (SSIM .697 to .819), confirming that explicit structural guidance is particularly valuable when stochastic sampling can induce morphological drift.
These improvements hold across both chromogen types and training set sizes differing by an order of magnitude, supporting generality of the proposed strategy.
Ki67 positivity error remains comparable across most architectures, while per-pixel MAE consistently improves, indicating that structural gains are largely preserved in marker quantification.

\subsection{Ablation Study}

Table~\ref{tab:ablation} compares four conditions per architecture on HNSCC~\cite{ghahremani2023ai}. The conditions are no prior (None), binary mask $M_\text{cell}$ (Binary), soft probability map $P_\text{soft}$ (Soft), and soft prior with variance-preserving loss (Soft+Var).
Both prior representations improve over the baseline, with the soft map consistently matching or outperforming the binary mask in SSIM and LPIPS, supporting the information-theoretic argument that continuous probabilities provide a richer conditioning signal.
The binary mask degrades LPIPS for Pix2Pix UNet despite improving SSIM (.175 vs .134), suggesting that hard boundaries can introduce perceptual artifacts even when structural alignment improves.
Adding variance-preserving loss yields the largest additional gain for Diffusion (SSIM .790 to .819, pMAE .028 to .021), the model most susceptible to texture smoothing, while providing marginal benefit for GAN and regression models where adversarial or reconstruction losses already preserve local contrast.

\subsection{Qualitative Analysis}

Figure~\ref{fig:qual} presents visual comparisons on DeepLIIF test cases across DAPI, Lap2, and Ki67 channels.
Baseline translators exhibit blurred nuclear boundaries and merged cells in cropped regions; the proposed method consistently reduces these artifacts, with the most pronounced gains in diffusion.
Ki67 channel outputs preserve punctate staining patterns more faithfully under the proposed method, maintaining clearer separation between positive and negative nuclei.

\section{Conclusion}
In this work, an architecture-agnostic soft prior conditioning strategy is presented that concatenates a continuous cell probability map from a pretrained nuclei segmentation model as a single additional input channel.
The proposed method consistently improves structural fidelity across adversarial, regression, and diffusion-based translators on two independent datasets without task-specific training.
The accompanying variance-preserving regularization complements the structural prior by maintaining local intensity heterogeneity, with the soft prior outperforming its binarized counterpart by retaining gradient-level boundary information.
These results highlight off-the-shelf nuclei priors as a practical route toward clinically reliable virtual staining.

\bibliographystyle{splncs04}
\bibliography{ref}

\end{document}